\documentstyle[11pt,colacl]{article}

\input{psfig}

\newcommand{\comment}[1]{}
\newcommand{\order}[1]{{\cal O}(#1)}

\newcommand{\myterm}{\Sigma}
\newcommand{\nonterm}{N}

\newcommand{\ep}{\varepsilon}

\newcommand{\mylabel}{{\it label}}
\newcommand{\mychild}{{\it children}}
\newcommand{\myroot}[1]{R_{{#1}}}
\newcommand{\myfoot}[1]{F_{{#1}}}
\newcommand{\mywrap}[1]{W_{{#1}}}
\newcommand{\myadj}{{\it Adj}}
\newcommand{\mysubst}{{\it Sbst}}
\newcommand{\mynil}{{\rm nil}}

\newcommand{\mybot}{B}
\newcommand{\myhalf}{M}
\newcommand{\mytop}{T}

\newcommand{\size}[1]{\mbox{$\mid \! {#1} \! \mid$}}

\title { Restrictions on Tree Adjoining Languages }
\author
 {
   Giorgio Satta \\
   Dip.~di Elettronica e Informatica \\
   Universit\`a di Padova \\
   35131 Padova, Italy \\
   {\tt satta@dei.unipd.it} \\
 \And
   William Schuler \\
   Computer and Information Science Dept. \\
   University of Pennsylvania \\
   Philadelphia, PA 19103 \\
   {\tt schuler@linc.cis.upenn.edu} \\
 }

\begin{document}


\maketitle

\begin{abstract}
   Several methods are known for parsing languages generated by
   Tree Adjoining Grammars (TAGs) in $\order{n^6}$ worst case running time.
   In this paper we investigate which restrictions on TAGs and
   TAG derivations are needed in order to lower this $\order{n^6}$ time
   complexity, without introducing large runtime constants, and
   without losing any of the generative power needed to capture
   the syntactic constructions in natural language that can be
   handled by unrestricted TAGs.  In particular, we describe an
   algorithm for parsing a strict subclass of TAG in $\order{n^5}$,
   and attempt to show that this subclass retains
   enough generative power to make it useful in the general case.
\end{abstract}




\section{Introduction}

Several methods are known that can parse languages
generated by Tree Adjoining Grammars (TAGs) in worst case time $\order{n^6}$, 
where $n$ is the length of the input string 
(see~\cite{schabes91} and references therein). 
Although asymptotically faster methods can be constructed,
as discussed in~\cite{rajasekaran95}, these methods are not of 
practical interest, due to large hidden constants. 
More generally, in~\cite{satta94} it has been argued 
that methods for TAG parsing running in time asymptotically faster 
than $\order{n^6}$ are unlikely to have small hidden constants. 

A careful inspection of the proof provided in~\cite{satta94}
reveals that the source of the claimed computational complexity 
of TAG parsing resides in the fact that 
auxiliary trees can get adjunctions at (at least) two distinct
nodes in their spine (the path 
connecting the root and the foot nodes). 
The question then arises of whether the bound of two 
is tight.
More generally, in this paper we investigate which 
restrictions on TAGs are needed in order to 
lower the $\order{n^6}$ time complexity, still retaining
the generative power that is needed to capture the
syntactic constructions of natural language that 
unrestricted TAGs can handle.
The contribution of this paper is twofold:
\begin{itemize} 
\item 
We define a strict subclass of TAG 
where adjunction of so-called wrapping trees at the spine 
is restricted to take place at no more than one distinct node. 
We show that in this case the parsing problem for TAG can
be solved in worst case time $\order{n^5}$. 
\item 
We provide evidence that the proposed subclass 
still captures the vast majority of TAG analyses that
have been currently proposed for the syntax 
of English and of several other languages. 
\end{itemize} 

Several restrictions on the adjunction operation for TAG have been proposed
in the literature~\cite{schabeswaters93,schabeswaters95} \cite{rogers94}.  
Differently from here, in all those works the main goal was one of characterizing, 
through the adjunction operation, the set of trees that can be generated by a
context-free grammar (CFG). 
For the sake of critical comparison, 
we discuss some common syntactic constructions 
found in current natural language TAG analyses, 
that can be captured by our proposal but fall 
outside of the restrictions mentioned above. 



\section{Overview}
\label{s:idea}

We introduce here the subclass of TAG that 
we investigate in this paper, and briefly compare
it with other proposals in the literature. 

A TAG is a tuple $G = (\nonterm, \myterm, I, A, S)$, 
where $\nonterm$, $\myterm$ are the finite sets of nonterminal 
and terminal symbols, respectively, $I$, $A$ are the finite 
sets of initial and auxiliary trees, respectively,
and $S \in \nonterm$ is the initial symbol. 
Trees in $I \cup A$ are also called elementary trees. 
The reader is referred to~\cite{joshi85} for the 
definitions of tree adjunction, tree substitution, 
and language derived by a TAG. 

The {\bf spine} of an auxiliary tree is 
the (unique) path that connects the root and the foot node. 
An auxiliary tree $\beta$ is called a {\bf right} ({\bf left}) tree if
(i) the leftmost (rightmost, resp.) leaf in $\beta$ is the foot node; and
(ii) the spine of $\beta$ contains only the root and the foot nodes.  
An auxiliary tree which is neither left nor right
is called a {\bf wrapping} tree.%
\footnote{The above names are also used in~\cite{schabeswaters95}
for slightly different kinds of trees.}


The {\bf TAG restriction} we propose is stated as followed:

\begin{enumerate}
\item 
At the spine of each wrapping tree, there is at most one node
that can host adjunction of a wrapping tree.  
This node is called a {\bf wrapping} node.  
\item
\label{i:lr}
At the spine of each left (right) tree, no wrapping tree can be 
adjoined and no adjunction constraints on right 
(left, resp.) auxiliary trees are found.  
\end{enumerate}

\noindent 
The above restriction does not in any way constrain
adjunction at nodes that are not in the spine of an auxiliary tree.  
Similarly, there is no restriction on the adjunction 
of left or right trees at the spines of wrapping trees.  

Our restriction is fundamentally different from those
in~\cite{schabeswaters93,schabeswaters95} and~\cite{rogers94},
in that we allow wrapping auxiliary trees to nest inside each
other an unbounded number of times, so long as they only adjoin
at one place in each others' spines.
Rogers, in contrast, restricts the nesting of wrapping
auxiliaries to a number of times
bounded by the size of the grammar, and Schabes and Waters forbid
wrapping auxiliaries altogether, at any node in the grammar.

We now focus on the recognition problem, and 
informally discuss the computational advantages that
arise in this task when a TAG obeys the above restriction. 
These ideas are formally developed in the next section. 
Most of the tabular methods for TAG recognition 
represent subtrees of derived trees, rooted at some node
$N$ and having the same span within the input string, 
by means of items of the form $\langle N, i, p, q, j \rangle$. 
In this notation $i$, $j$ are positions in the input 
spanned by $N$, and $p$, $q$ are positions spanned
by the foot node, in case $N$ belongs to the spine,
as we assume in the discussion below. 

    \begin{figure}[htbp]
    \centerline{\hbox{\psfig{figure=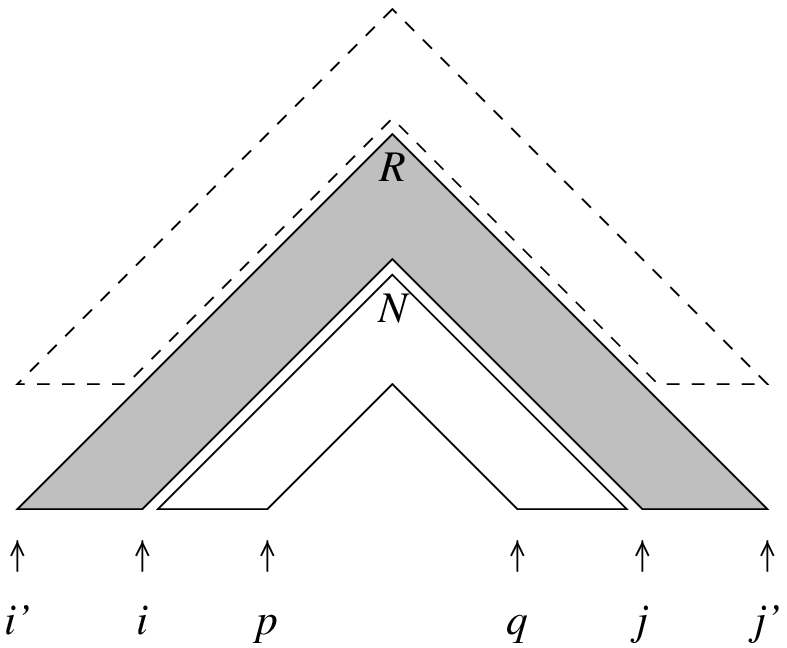,height=1.5in}}}
    \caption{$\order{n^6}$ wrapping adjunction step.}
    \label{f:adj}
    \end{figure}

The most time expensive step in TAG recognition is the one that 
deals with adjunction.  When we adjoin at $N$ a derived auxiliary tree 
rooted at some node $R$, we have to combine together two items 
$\langle R, i', i, j, j' \rangle$ and $\langle N, i, p, q, j \rangle$.
This is shown in Figure~\ref{f:adj}. 
This step involves six different indices that could range 
over any position in the input, and thus has a time
cost of $\order{n^6}$.  

Let us now consider adjunction of wrapping trees, 
and leave aside left and right trees for the moment.  
Assume that no adjunction has been performed in the portion 
of the spine below $N$. Then none of the trees adjoined below 
$N$ will simultaneously affect the portions of the tree yield
to the left  and to the right of the foot node. 
In this case we can safely split the tree yield and 
represent item $\langle N, i, p, q, j \rangle$ by means of two items 
of a new kind, $\langle N_{\it left}, i, p \rangle$ and 
$\langle N_{\it right}, q, j \rangle$.  
The adjunction step can now be performed 
by means of two successive steps.
The first step combines $\langle R, i', i, j, j' \rangle$ and 
$\langle N_{\it left}, i, p \rangle$, producing
a new intermediate item~$I$.   The second step 
combines $I$ and $\langle N_{\it right}, q, j \rangle$,
producing the desired result.
In this way the  time cost is reduced to $\order{n^5}$. 

    \begin{figure}[htbp]
    \centerline{\hbox{\psfig{figure=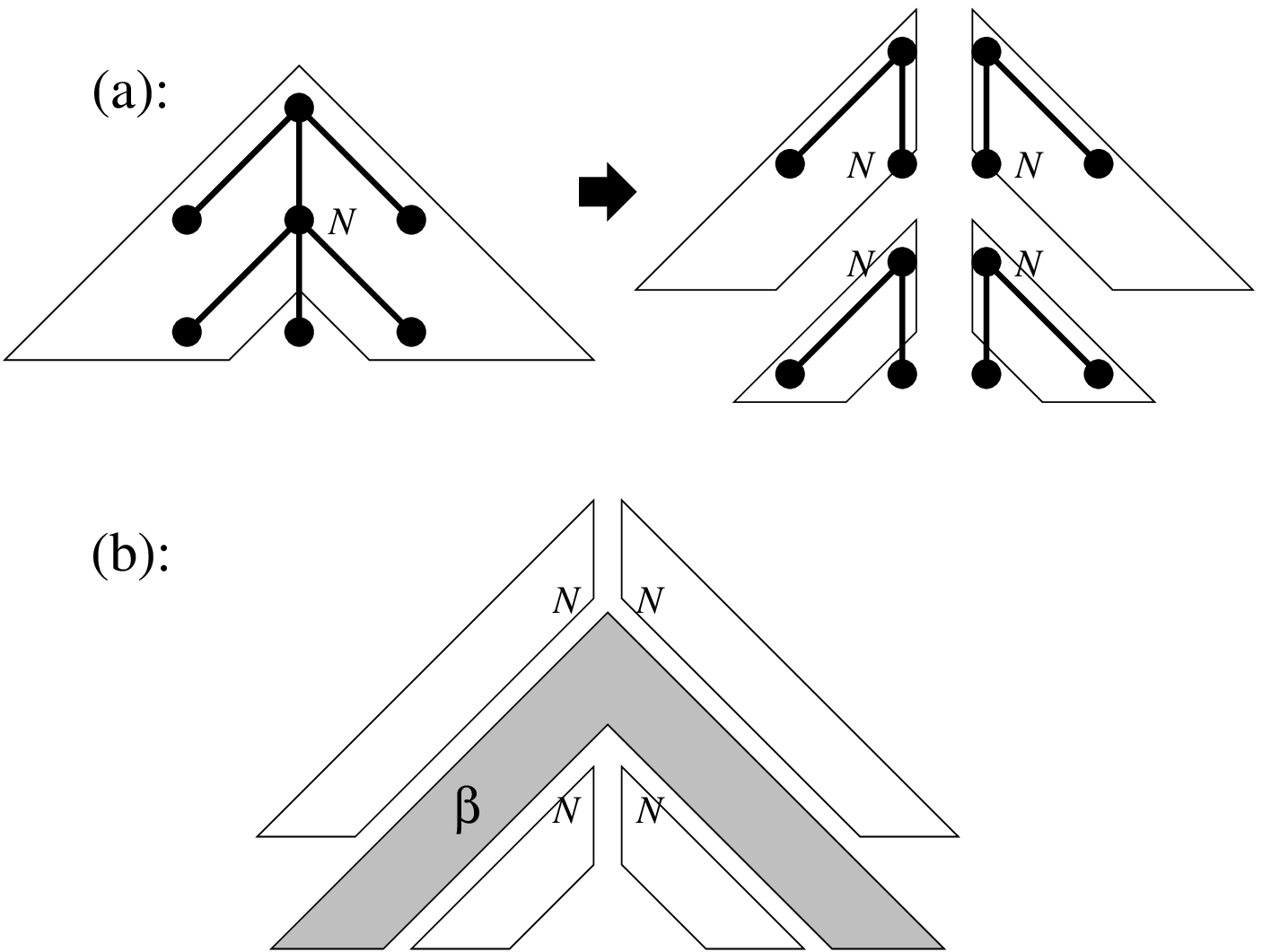,height=2.5in}}}
    \caption{$\order{n^5}$ wrapping adjunction step.}
    \label{f:split}
    \end{figure}

It is not difficult to see that the above reasoning also applies 
in cases where no adjunction has been performed at the portion 
of the spine above $N$.  
This suggests that, when processing a TAG that obeys the restriction 
introduced above, we can always `split' each wrapping tree into four parts 
at the wrapping node $N$, since $N$ is the only site in the spine 
that can host adjunction (see Figure~\ref{f:split}(a)).  
Adjunction of a wrapping tree $\beta$ at $N$ 
can then be simulated by four steps, executed one after the other. 
Each step composes the item resulting from the application of 
the previous step with an item representing one of the four parts 
of the wrapping tree (see Figure~\ref{f:split}(b)).  

We now consider adjunction involving left and right trees, 
and show that a similar splitting along the spine can be 
performed.
Assume that $\gamma$ is a derived auxiliary tree, 
obtained by adjoining several left and right trees one at the 
spine of the other.  
Let $x$ and $y$ be the part of the yield of $\gamma$
to the left and right, respectively, of the foot node. 
{}From the definition of left and right trees, we have that 
the nodes in the spine of $\gamma$ have all the same
nonterminal label.  
Also, from condition~\ref{i:lr} in the above 
restriction we have that the  
left trees adjoined in $\gamma$ do not constrain 
in any way the right trees adjoined in $\gamma$.  
Then the following derivation can always be performed. 
We adjoin all the left trees, each one at the spine of the
other, in such a way that the resulting tree 
$\gamma_{\it left}$ has yield $x$.  
Similarly, we adjoining all the right trees, one at the spine of the
other, in such a way that the yield of the resulting tree 
$\gamma_{\it right}$ is $y$.  
Finally, we adjoin $\gamma_{\it right}$
at the root of $\gamma_{\it left}$, obtaining a derived tree having
the same yield as $\gamma$. 

{}From the above observations it directly follows that 
we can always recognize the yield of $\gamma$
by independently recognizing $\gamma_{\it left}$
and $\gamma_{\it right}$.   
Most important, $\gamma_{\it left}$ and $\gamma_{\it right}$ 
can be represented by means of items 
$\langle R_{\it left}, i, p \rangle$ and $\langle R_{\it right}, q, j \rangle$.  
As before, the adjunction of tree $\gamma$ at some subtree represented by an 
item $I$ can be recognized by means of two successive steps, 
one combining $I$ with $\langle R_{\it left}, i, p \rangle$ at its left, 
resulting in an intermediate item $I'$, and the second 
combining $I'$ with $\langle R_{\it right}, q, j \rangle$ at its right, 
obtaining the desired result. 



\section{Recognition}
\label{s:algo}

This section presents the main result of the paper. 
We provide an algorithm for the recognition of languages 
generated by the subclass of TAGs introduced 
in the previous section, and show that the worst case running time 
is $\order{n^5}$, where $n$ is the length of the input string. 
To simplify the presentation, we assume the following conditions
throughout this section: 
first, that elementary trees are binary (no more than two children at each node) 
and no leaf node is labeled by $\ep$; and
second, that there is always a wrapping node in each wrapping tree, 
and it differs from the foot and the root node. 
This is without any loss of generality.

\subsection{Grammar transformation}
\label{ss:transf}

Let $G = (\nonterm, \myterm, I, A)$ 
be a TAG obeying the restrictions of Section~\ref{s:idea}. 
We first transform $A$ into a new set 
of auxiliary trees $A'$ that will be processed by our method. 
The root and foot nodes of a tree $\beta$ are denoted
$\myroot{\beta}$ and $\myfoot{\beta}$, respectively.  
The wrapping node (as defined in Section~\ref{s:idea}) 
of $\beta$ is denoted $\mywrap{\beta}$. 

Each left (right) tree $\beta$ in $A$ is inserted in $A'$
and is called $\beta_L$ ($\beta_R$). 
Let $\beta$ be a wrapping tree in $A$.
We split $\beta$ into four 
auxiliary trees, as informally described in Section~\ref{s:idea}. 
Let $\beta_D$ be the subtree of $\beta$ rooted at $\mywrap{\beta}$.
We call $\beta_U$ the tree obtained from $\beta$
by removing every descendant of $\mywrap{\beta}$
(and the corresponding arcs).  
We remove every node to the right (left) of the spine of $\beta_D$
and call $\beta_{LD}$ ($\beta_{RD}$) the resulting tree.
Similarly, 
we remove every node to the right (left) of the spine of $\beta_U$
and call $\beta_{LU}$ ($\beta_{RU}$) the resulting tree.
We set $\myfoot{\beta_{LD}}$ and $\myfoot{\beta_{RD}}$ equal to $\myfoot{\beta}$,
and set $\myfoot{\beta_{LU}}$ and $\myfoot{\beta_{RU}}$ equal to $\mywrap{\beta}$.
Trees $\beta_{LU}$, $\beta_{RU}$, $\beta_{LD}$, and $\beta_{RD}$ are
inserted in $A'$ for every wrapping tree $\beta$ in $A$.

Each tree in $A'$ inherits at its nodes the adjunction constraints
specified in $G$.  In addition, we impose the following constraints:
\begin{itemize}
\item 
only trees $\beta_L$ can be adjoined at the spine of 
trees $\beta_{LD}, \beta_{LU}$;
\item 
only trees $\beta_R$ can be adjoined at the spine of 
trees $\beta_{RD}, \beta_{RU}$;
\item
no adjunction can be performed at 
nodes $\myfoot{\beta_{LU}}, \myfoot{\beta_{RU}}$. 
\end{itemize}

\subsection{The algorithm}

The algorithm below is a tabular method that works bottom up on
derivation trees.  
Following~\cite{shieberetal95}, we specify the algorithm using inference rules. 
(The specification has been optimized for presentation simplicity,
not for computational efficiency.)

Symbols $N,P,Q$ denote nodes of trees in $A'$ (including foot and root), 
$\alpha$ denotes initial trees and $\beta$ denotes auxiliary trees.
Symbol $\mylabel(N)$ is the label of $N$ and 
$\mychild(N)$ is a string denoting all children of $N$ from 
left to right 
($\mychild(N)$ is undefined if $N$ is a leaf).  
We write $\alpha \in \mysubst(N)$ if $\alpha$ can be substituted at $N$.
We write $\beta \in \myadj(N)$ if $\beta$ can be adjoined at $N$, 
and $\mynil \in \myadj(N)$ if adjunction at $N$ is optional.   

We use two kind of items:
\begin{itemize}
\item 
Item $\langle N^{X}, i,j \rangle$, 
$X \in \{\mybot, \myhalf, \mytop\}$, denotes a subtree 
rooted at $N$ and spanning the portion of the input from 
$i$ to $j$.  Note that two input positions are sufficient, 
since trees in $A'$ always have their foot node at the 
position of the leftmost or rightmost leaf. 
We have $X=\mybot$ if $N$ has not yet been processed 
for adjunction, $X=\myhalf$ if $N$ has been processed only 
for adjunction of trees $\beta_L$, and $X=\mytop$ if $N$ has already 
been processed for adjunction.  
\item
Item $\langle \beta, i,p,q,j \rangle$ denotes a wrapping 
tree $\beta$ (in $A$) with $\myroot{\beta}$ spanning 
the portion of the input from 
$i$ to $j$ and with $\myfoot{\beta}$ spanning 
the portion of the input from $p$ to $q$.  
In place of $\beta$ we might use symbols
$[\beta, LD]$, $[\beta, RD]$ and $[\beta, RU]$ to
denote the temporary results of recognizing 
the adjunction of some wrapping tree at $\mywrap{\beta}$. 
\end{itemize}

\vspace{5pt}
\noindent
{\bf Algorithm}. \hspace{0.6em}
Let $G$ be a TAG with the restrictions of Section~\ref{s:idea},
and let $A'$ be the associated set of auxiliary trees defined
as in section~\ref{ss:transf}.  
Let $a_1 a_2 \cdots a_n$, $n \geq 1$, be an input string. 
The algorithm accepts the input iff some item 
$\langle R^{\mytop}_{\alpha}, 0, n \rangle$
can be inferred for some $\alpha \in I$. 

\vspace{2mm}

\noindent {\bf Step 1} This step recognizes subtrees with root $N$ from 
subtrees with roots in $\mychild(N)$.  

\small

\(
\begin{array}{c}
 \mbox{  } \\
 \hline
 \langle N^{\mytop},i-1,i \rangle
\end{array}, 
\; \mylabel(N) = a_i;
\)

\(
\begin{array}{c}
 \mbox{  } \\
 \hline
 \langle \myfoot{\beta}^{\mybot},i,i \rangle
\end{array}, 
\; \beta \in A', \;  0 \leq i \leq n; 
\)

\(
\begin{array}{c}
 \langle \myroot{\alpha}^{\mytop},i,j \rangle \\
 \hline
 \langle N^{\mytop},i,j \rangle
\end{array}, 
\; \alpha \in \mysubst(N); 	
\)

\(
\begin{array}{cc}
 \langle P^{\mytop},i,k \rangle & 
 \langle Q^{\mytop},k,j \rangle \\
 \hline
 \multicolumn{2}{c}{\langle N^{\mybot},i,j \rangle} \\
\end{array}, 
\; \mychild(N)~=~PQ; 	
\)

\(
\begin{array}{c}
 \langle P^{\mytop},i,j \rangle \\
 \hline
 \langle N^{\mybot},i,j \rangle
\end{array}, 
\; \mychild(N) = P. 
\) 

\normalsize

\vspace{2mm}

\noindent {\bf Step 2} This step recognizes the adjunction of wrapping
trees at wrapping nodes.   We recognize the tree
hosting adjunction by composing its four `chunks', 
represented by auxiliary trees 
$\beta_{LD}$, $\beta_{RD}$, $\beta_{RU}$ and 
$\beta_{LU}$ in $A'$, around the wrapped tree.  

\small

\(
\begin{array}{cc}
 \! \langle R_{\beta_{LD}}^{\mybot}, k,p \rangle \! &
 \! \langle \beta',i,k,q,j \rangle \! \\
 \hline 
 \multicolumn{2}{c}{\langle [\beta,LD],i,p,q,j \rangle} 
\end{array},
\beta' \in \myadj(\mywrap{\beta}), p < q; 
\) 

\(
\begin{array}{cc}
 \langle R_{\beta_{RD}}^{\mybot}, q,k \rangle & 
 \langle [\beta,LD],i,p,k,j \rangle \\
 \hline
 \multicolumn{2}{c}{\langle [\beta,RD],i,p,q,j \rangle} 
\end{array},
\; p<q;
\) 

\(
\begin{array}{cc}
 \langle R_{\beta_{RU}}^{\mytop}, k,j \rangle & 
 \langle [\beta,RD],i,p,q,k \rangle \\
 \hline
 \multicolumn{2}{c}{\langle [\beta,RU], i,p,q,j \rangle} 
\end{array};
\) 

\(
\begin{array}{cc}
 \langle R_{\beta_{LU}}^{\mytop}, i,k \rangle & 
 \langle [\beta,RU],k,p,q,j \rangle \\
 \hline
 \multicolumn{2}{c}{\langle \beta, i,p,q,j \rangle} 
\end{array}; 
\) 

\(
\begin{array}{cc}
 \! \langle R_{\beta_{LD}}^{\mytop}, i,p \rangle \! & 
 \! \langle R_{\beta_{RD}}^{\mytop}, q,j \rangle \! \\
 \hline
 \multicolumn{2}{c}{\langle [\beta,RD], i,p,q,j \rangle} 
\end{array}, 
\mynil \in \myadj(\mywrap{\beta}), p < q. 
\) 

\normalsize

\vspace{2mm}

\noindent {\bf Step 3} This step recognizes all remaining cases of adjunction. 

\small

\(
\begin{array}{cc}
 \! \langle R_{\beta_{L}}^{\mytop},i,k \rangle \! &
 \! \langle N^{\mybot},k,j \rangle \! \\  
 \hline 
 \multicolumn{2}{c}{\langle N^{X},i,j \rangle}
\end{array},
\beta\!\in\!\myadj(N),X\!\in\!\{\myhalf,\mytop\};
\)

\(
\begin{array}{cc}
 \! \langle N^{X}\!,i,k \rangle \! &
 \! \langle R_{\beta_{R}}^{\mytop},k,j \rangle \! \\  
 \hline 
 \multicolumn{2}{c}{\langle N^{\mytop},i,j \rangle}
\end{array},
\beta\!\in\!\myadj(N),X\!\in\!\{\mybot,\myhalf\};
\)

\(
\begin{array}{c}
 \langle N^{\mybot}, i,j \rangle  \\
 \hline
 \langle N^{\mytop}, i,j \rangle 
\end{array},
\; \mynil \in \myadj(N); 
\)

\(
\begin{array}{cc}
 \langle N^{\mybot}, p,q \rangle & 
 \langle \beta, i,p,q,j \rangle \\  
 \hline 
 \multicolumn{2}{c}{\langle N^{\mytop}, i,j \rangle}
\end{array},
\; \beta \in \myadj(N).
\)

\normalsize

\vspace{2mm}

Due to restrictions on space, we merely claim the
correctness of the above algorithm.
We now establish its worst case time complexity with
respect to the input string length $n$. 
We need to consider the maximum number $d$ of input positions
appearing in the antecedent of an inference rule. 
In fact, in the worst case
we will have to execute a number of different evaluations
of each inference rule which is proportional to $n^d$, 
and each evaluation can be carried out in an 
amount of time independent of $n$. 
It is easy to establish that Step~1 can 
be executed in time $\order{n^3}$ and that Step~3 can 
be executed in time $\order{n^4}$. 
Adjunction at wrapping nodes performed at Step~2 is the most 
expensive operation, requiring an amount of time $\order{n^5}$.
This is also the time complexity of our algorithm.


\section{Linguistic Relevance}
\label{s:ling}

In this section we will attempt to show that the restricted formalism
presented in Section~\ref{s:idea}
retains enough generative power to make it useful in the general case.

\subsection{Athematic and Complement Trees}

We begin by introducing the distinction between athematic 
auxiliary trees and complement auxiliary trees~\cite{kroch89},
which are meant to exhaustively characterize the auxiliary trees
used in any natural language TAG grammar.%
\footnote{The same linguistic distinction is used in the conception
of `modifier' and `predicative' trees~\cite{schabesshieber94}, but
Schabes and Shieber give the trees special properties in the calculation
of derivation structures, which we do not.}
An {\bf athematic} auxiliary tree does not subcategorize for or assign
a thematic role to its foot node, so the head of the foot node becomes
the head of the phrase at the root.
The structure of an athematic auxiliary tree may thus be described as:
    \begin{equation}
    X^{n} \rightarrow X^{n} \cdots (Y^{{\it max}}) \cdots,
    \end{equation} 
\noindent where $X^{n}$ is any projection 
of category $X$, $Y^{{\it max}}$ is the
maximal projection of $Y$, and the order of the constituents is
variable.%
\footnote{The~CFG-like notation~is~taken~directly from~\cite{kroch89},
where it is used to specify labels at the root and frontier nodes of a tree
without placing constraints on the internal structure.}
A {\bf complement} auxiliary tree, on the other hand, introduces a lexical head
that subcategorizes for the tree's foot node and assigns it a thematic role.
The structure of a complement auxiliary tree may be described as:
    \begin{equation}
    X^{{\it max}} \rightarrow \cdots Y^{0} \cdots X^{{\it max}} \cdots,
    \end{equation}
\noindent where $X^{{\it max}}$ is the maximal projection of some category $X$,
and $Y^{0}$ is the lexical projection of some category $Y$, whose maximal
projection dominates $X^{{\it max}}$.

{}From this we make the following observations:

\begin{enumerate}
  \item Because it does not assign a theta role to its foot node, an
        athematic auxiliary tree may adjoin at any projection of a
        category, which we take to designate any adjunction site in
        a host elementary tree.
  \item Because it does assign a theta role to its foot node, a
        complement auxiliary tree may only adjoin at a certain
        `complement' adjunction site in a host elementary tree,
        which must at least be a maximal projection of a lexical category.
  \item The foot node of an athematic auxiliary tree is dominated
        only by the root, with no intervening nodes, so it falls outside
        of the maximal projection of the head.
  \item The foot node of a complement auxiliary tree is dominated
        by the maximal projection of the head, which may also dominate
        other arguments on either side of the foot.
\end{enumerate}

To this we now add the assumption that each auxiliary tree can have
only one complement adjunction site projecting from $Y^{0}$,
where $Y^{0}$ is the lexical category that projects $Y^{{\it max}}$.
This is justified in order to prevent projections of $Y^{0}$ from receiving
more than one theta role from complement adjuncts, which would violate the
underlying theta criterion in Government and Binding Theory~\cite{chomsky81}.%
We also assume that an auxiliary tree can not have complement
adjunction sites on its spine projecting from lexical heads other than $Y^{0}$,
in order to preserve the minimality of elementary trees~\cite{kroch89,frank92}.
Thus there can be no more than one complement adjunction site on
the spine of any complement auxiliary tree, and no complement adjunction
site on the spine of any athematic auxiliary tree, since the foot node
of an athematic tree lies outside of the maximal projection of the head.%
\footnote{It is important to note that, in order to satisfy the theta
criterion and minimality, we need only constrain the number of complement
adjunctions -- not the number of complement adjunction sites -- on the spine
of an auxiliary tree.
Although this would remain within the power of our formalism, we prefer to
use constraints expressed in terms of adjunction sites, as we did in
Section~\ref{s:idea}, because it provides a restriction on elementary trees,
rather than on derivations.}

Based on observations 3 and 4, we can further specify that only complement
trees may wrap, because the foot node of an athematic tree lies outside of
the maximal projection of the head, below which all of its subcategories
must attach.%
\footnote { Except in the case of raising, discussed below. }
In this manner, we can insure that only one wrapping tree (the
complement auxiliary) can adjoin into the spine of a wrapping (complement)
auxiliary, and only athematic auxiliaries (which must be left/right trees)
can adjoin elsewhere,
fulfilling our TAG restriction in Section~\ref{s:idea}.%

\subsection{Possible Extensions}
\label{ss:extensions}

\begin{figure}[htbp]
    \centerline{\hbox{\psfig{figure=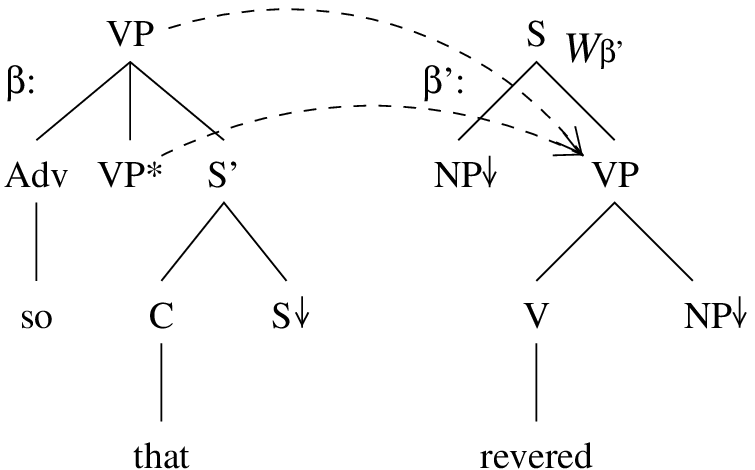,height=1.2in}}}
    \caption{Wrapping athematic tree.}
    \label{f:sothat}
    \end{figure}

We may want to weaken our definition to include wrapping athematic
auxiliaries, in order to account for modifiers with raised
heads or complements as in Figure~\ref{f:sothat}: ``They so revered him
that they built a statue in his honor.''
This can be done within the above algorithm as long as the athematic
trees do not wrap productively (that is as long as they cannot be
adjoined one at the spine of the other) by splitting the athematic
auxiliary tree down the spine and treating the two fragments as
tree-local multi-components, which can be simulated with non-recursive
features~(Hockey and Srinivas, 1993)\nocite{hockeysrini93}.

\noindent Since the added features are non-recursive, this extension
would not alter the $\order{n^5}$ result reported in Section~\ref{s:algo}.

\subsection{Comparison of Coverage}

\begin{figure}[htbp]
    \centerline{\hbox{\psfig{figure=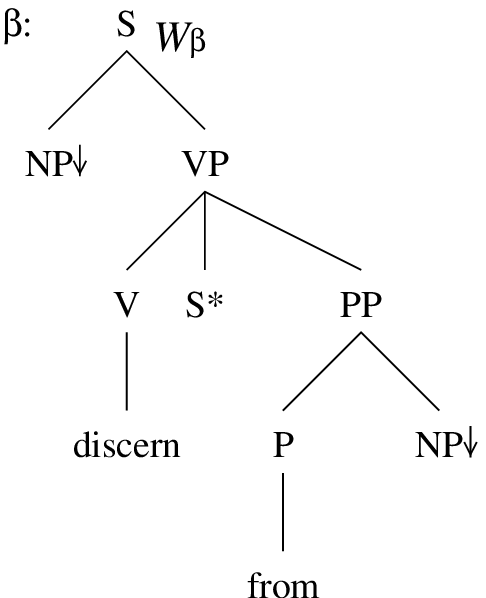,height=1.7in}}}
    \caption{Wrapping complement tree.}
    \label{f:discerned}
    \end{figure}

\begin{figure}[htbp]
    \centerline{\hbox{\psfig{figure=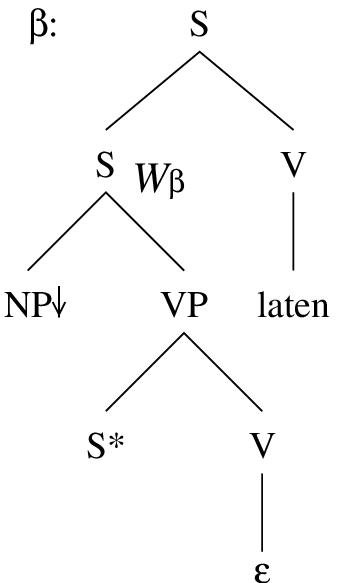,height=1.7in}}}
    \caption{Verb-raising tree in Dutch.}
    \label{f:dutch}
    \end{figure}

In contrast to the formalisms of Schabes and Waters
\cite{schabeswaters93,schabeswaters95}, our restriction allows
wrapping complement auxiliaries as in
Figure~\ref{f:discerned}~\cite{schabeswaters95}.
Although it is difficult to find examples in English which are excluded
by Rogers' regular form restriction~\cite{rogers94}, we can
cite verb-raised complement auxiliary trees in Dutch as in Figure
\ref{f:dutch}~\cite{krochsantorini91}.
Trees with this structure may adjoin into each others' internal spine
nodes an unbounded number of times, in violation of Rogers' definition
of regular form adjunction, but within our criteria of wrapping
adjunction at only one node on the spine.



\section{Concluding remarks}
\label{s:final}

Our proposal is intended to contribute to the assessment of the
computational complexity of syntactic processing. 
We have introduced a strict subclass of TAGs having the generative
power that is needed to account for the syntactic constructions of
natural language that unrestricted TAGs can handle.
We have specified a method that recognizes the generated languages
in worst case time $\order{n^5}$, where $n$ is the length of the
input string. 
In order to account for the dependency on the 
input grammar $G$, let us define 
$\size{G} = \sum_{N} (1 + \size{\myadj(N)})$, where $N$ ranges
over the set of all nodes of the elementary trees. 
It is not difficult to see that the running
time of our method is proportional to $\size{G}$. 

Our method works as a recognizer.  As for many other tabular 
methods for TAG recognition, we can devise simple procedures 
in order to obtain a derived tree associated with an accepted string. 
To this end, we must be able to `interleave' adjunctions
of left and right trees, that are always kept separate
by our recognizer.

The average case time complexity of our method should surpass its
worst case time performance, as is the case for many other tabular
algorithms for TAG recognition. 
In a more applicative perspective, then, 
the question arises of whether there is any gain in 
using an algorithm that is unable to recognize more
than one wrapping adjunction at each spine, as opposed
to using an unrestricted TAG algorithm.  
As we have tried to argue in Section~\ref{s:ling}, 
it seems that standard syntactic constructions do not exploit
multiple wrapping adjunctions at a single spine. 
Nevertheless, the local ambiguity of natural language,
as well as cases of ill-formed input, could always produce cases in which 
such expensive analyses are attempted by an unrestricted algorithm.
In this perspective, then, we conjecture that 
having the single-wrapping-adjunction restriction
embedded into the recognizer would improve processing efficiency
in the average case. 
Of course, more experimental work would be needed in order
to evaluate such a conjecture, which we leave for future work. 


\section*{Acknowledgments}

Part of this research was done while the first author 
was visiting the Institute for Research in Cognitive Science, 
University of Pennsylvania.
The first author was supported by NSF grant SBR8920230. 
The second author was supported by
U.S. Army Research Office Contract No. DAAH04-94G-0426.
The authors would like to thank
 Christy Doran, 
 Aravind Joshi,
 Anthony Kroch, 
 Mark-Jan Nederhof,
 Marta Palmer,
 James Rogers and 
 Anoop Sarkar
for their help in this research.


\end{document}